\documentclass{article}
\usepackage{spconf,amsmath}
\usepackage{verbatim,booktabs}
\usepackage{todonotes}
\usepackage{textcomp}
\usepackage{graphicx}
\usepackage{subcaption}
\usepackage{url}

\setuptodonotes{inline}

\ninept

\title{Deploying a BERT-based Query-Title Relevance Classifier in a Production System: a View from the Trenches}
\name{Leonard Dahlmann, Tomer Lancewicki}
\address{eBay Inc.}
\begin{document}
%
\maketitle
\begin{abstract}
The Bidirectional Encoder Representations from Transformers (BERT) model has been radically improving the performance of many Natural Language Processing (NLP) tasks such as Text Classification and Named Entity Recognition (NER) applications. However, it is challenging to scale BERT for low-latency and high-throughput industrial use cases due to its enormous size. 
We successfully optimize a Query-Title Relevance (QTR) classifier for deployment via a compact model, which we name BERT Bidirectional Long Short-Term Memory (BertBiLSTM). The model is capable of inferring an input in at most 0.2ms on CPU.
BertBiLSTM exceeds the off-the-shelf BERT model's performance in terms of accuracy and efficiency for the aforementioned real-world production task. We achieve this result in two phases. First, we create a pre-trained model, called eBERT, which is the original BERT architecture trained with our unique item title corpus. We then fine-tune eBERT for the QTR task. Second, we train the BertBiLSTM model to mimic the eBERT model's performance through a process called Knowledge Distillation (KD) and show the effect of data augmentation to achieve the resembling goal. Experimental results show that the proposed model outperforms other compact and production-ready models.

\end{abstract}
\begin{keywords}
Knowledge Distillation,
Long Short-Term Memory (LSTM), 
Self-Attention, 
Pre-Training, 
Contextual Embedding 
\end{keywords}

\section{Introduction}
\label{sec:intro}
One of the most significant challenges data scientists face in Natural Language Processing (NLP) tasks is the scarcity of human-labeled data. However, to achieve state-of-the-art results, modern deep learning NLP architectures require a large amount of training data. 
A common approach to solve this problem is to use a pre-trained model that serves as the starting point for different but related tasks. The Bidirectional Encoder Representations from Transformers (BERT) model \cite{devlin2018bert} is a revolutionary pre-trained model that has helped the industry achieve remarkable results in many NLP tasks. By reusing the parameters of BERT, we can quickly fine-tune it for different use cases with a relatively small amount of training data, save significant amounts of training time and cost, and still achieve prominent results.

However, as the BERT-base model starts at 110 million parameters, it is challenging to scale it for low-latency and high-throughput use cases. From a hardware perspective, BERT inference with CPUs reduces deployment costs but does not meet the required throughput. Utilizing forefront hardware as GPUs tremendously increases the throughput but is not possible due to budget constraints. 
In this work, we tackle the problem of deploying a ranking algorithm, formulated as a Query-Title Relevance (QTR) binary classification problem \cite[Ch. 2.3]{liu2011learning}. The data consists of query-title pair inputs, labeled as relevant (1) or irrelevant (0).
 We show corpus statistics in Table \ref{tab:srdata} and some examples from the data in Table \ref{tab:srexamples}. Note that the
data set is imbalanced when the majority of pairs is labeled as relevant.

\begin{table*}[!h]
    \footnotesize
    \caption{Overview of our Query-Title Relevance (QTR) dataset.}
    \centering
    \begin{subtable}{.325\linewidth}
      \centering
        \caption{Corpus statistics.}
        \begin{tabular}{lrr}
         \toprule
         Corpus & Examples & Positive [\%] \\ \midrule
         Train & 764k & 78.5 \\
         Development & 52k & 79.9 \\
         Test & 155k & 80.1 \\
         \bottomrule
        \end{tabular}
        \label{tab:srdata}
    \end{subtable}%
    \begin{subtable}{.7\linewidth}
      \centering
        \caption{Examples.}
        \begin{tabular}{llr}
             \toprule
             Query & Title & Label \\ \midrule
             sony a55 & Sony Alpha SLT-A55 Digital Camera Body & 1 \\ 
             linksys cm3024 modem & Linksys High Speed 3.0 24x8 Cable Modem (CM3024) & 1 \\
             star war rebels bluray & Star wars rebels complete season 4 DVD & 0 \\
             \bottomrule
        \end{tabular}
        \label{tab:srexamples}
    \end{subtable} 
\end{table*}

We have found that a fine-tuned BERT model for the QTR classification task gives superior performances relative to our current in-production solutions. However, due to the aforementioned issues, we face the challenge of deploying the BERT-based classifier as part of our production system.
To overcome deployment challenges, we exploit Knowledge Distillation (KD) \cite{hinton2015distilling}, a compression method where a relatively simple model, called "student", is trained to fit the output distribution of a well trained "teacher" model as pseudo labels, instead of the original ground truth labels. KD proved to be effective in many domains, such as Sequence-to-Sequence (Seq2Seq) models for Large Vocabulary Continuous Speech Recognition (LVCSR) \cite{distill_icassp1}, Natural Language Processing (NLP) tasks \cite{tang2019natural}, and low-resolution visual recognition problem \cite{distill_icassp2}. 
\begin{table}[t]
    \footnotesize
    \centering
    \caption{Average inference latency of various models for the QTR task with a batch of size 1 for an input length of 22 on CPU. $L$ refers to the number of encoder layers and $d_\text{h}$ to the hidden size. For BERT layers, $d_\text{ff}$ is the intermediate size and $h$ the number of attention heads. }
    \begin{tabular}{lllllr}
    \toprule
    Model & $L$ & $d_\text{h}$ & $d_\text{ff}$ & $h$ & Latency [ms] \\ \midrule
    BERT-base \cite{devlin2018bert} & 12 & 768 & 3072 & 12 & 15.9 \\ 
    DistilBERT \cite{sanh2019distilbert} & 6 & 768 & 3072 & 12 & 7.95 \\ 
    TinyBERT \cite{jiao2019tinybert} & 4 & 312 & 1200 & 4 & 0.935 \\
    \midrule
    BiLSTM \cite{tang2019distilling} & 1 & 128 & - & - & 0.057 \\
    BiLSTM \cite{tang2019distilling} & 1 & 300 & - & - & 0.116 \\
    BERT-student & 1 & 128 & 128 & 4 & 0.055 \\
    BERT-student & 1 & 300 & 300 & 4 & 0.143 \\
    \bottomrule
    \end{tabular}
    \label{tab:speed}
\end{table}

Several recent works have proposed small student architectures for KD, with BERT as the teacher model. They differ in the phases in which KD is applied. DistilBERT \cite{sanh2019distilbert} and MobileBERT \cite{sun2020mobilebert} train smaller general-purpose BERT models by applying KD only in pre-training. TinyBERT \cite{jiao2019tinybert}
apply KD both in pre-training and fine-tuning. Finally, \cite{sun2019patient} use task-specific KD by training shallow BERT students in a multi-layer approach, while \cite{tang2019distilling} train task-specific single-layer BiLSTMs with KD and incorporate augmented training data. In this work, we follow the latter approach of applying KD only in fine-tuning. We have implemented the aforementioned architectures in our in-house CPU inference toolkit, which relies on oneDNN \cite{onednn} and FBGEMM \cite{fbgemm} for highly-optimized inference, exploiting the AVX-512 VNNI instruction set \cite{avx512}.
As shown in Table \ref{tab:speed}, even TinyBERT \cite{jiao2019tinybert} would be too slow for our use-case, for which inference latency must not exceed 0.2ms. We therefore restrict ourselves to student models with at most three BERT or BiLSTM layers and a hidden size of at most $d_\text{h}=d_\text{ff}=300$. We refer to BERT models satisfying this constraint as BERT-student.

We conjecture that BiLSTMs and self-attention have complementary strengths, which become apparent in small models: BiLSTMs can capture sequential relations, while self-attention is able to perform cross-checking between query and title tokens. Motivated by this, we show that we can improve over \cite{tang2019distilling} for QTR by training deeper but more narrow models, as well as cascading BERT-student and BiLSTM layers in the proposed BertBiLSTM model. Combinations of RNNs and Transformers \cite{vaswani2017attention} have been used before, e.g. for deep models in machine translation \cite{chen2018best}, speech recognition \cite{zeyer2019comparison} and text classification \cite{vlad2019sentence}. To the best of our knowledge, we are the first to apply such a hybrid model for the purpose of training small and production-oriented models with KD from pre-trained BERT models.

Along the path to achieving our goal of deploying a QTR classifier in production, we made the following contributions:
\begin{enumerate}
  \item We pre-train BERT from scratch with additional e-commerce data and refer to this model as eBERT.
  Although training BERT from scratch is expensive in comparison to using the off-the-shelf BERT model, it is worth the efforts as our results in Table \ref{tab:model_overview} show.
  \item We compare the performance of various BiLSTM and BERT-student models on the QTR task, and show that deeper but more narrow models outperform the shallow variants, while still satisfying our requirement for low latency.
  \item We introduce the hybrid student architecture BertBiLSTM, which composes a BiLSTM over position-wise BERT-student outputs and thereby outperforms the pure BiLSTM and BERT-student encoders on QTR. 
\end{enumerate}

The rest of the paper is organized as follows: 
In Section 2, we present our KD procedure to solve the QTR classification problem, and our proposed BertBiLSTM student model. We discuss experimental results in Section 3. Conclusions are summarized in Section 4.

\section{Knowledge Distillation for Query-Title Relevance task}
\label{sectionKD}
Knowledge distillation (KD) is a compression method \cite{hinton2015distilling} where a small "student" model is trained to fit the output distribution of a large "teacher" model as pseudo labels, instead of the original ground truth labels. In KD training, the teacher produces softmax probabilities (soft targets), which reveal the teacher’s confidences for what classes it predicts given an input. With this additional knowledge, the student can model the data distribution better than learning directly from the ground truth labels consisting of one-hot vectors (hard targets). 
In this section we provide the teacher and student architectures. 
\subsection{Teacher Architecture}
\label{sec:teacher}
We present our teacher, denoted as eBERT, which is the original BERT model \cite{devlin2018bert} with the following intensification: 
As in the original BERT paper \cite{devlin2018bert}, we pre-train the eBERT model with the BooksCorpus (800M words) \cite{zhu2015aligning} and English Wikipedia (2.5B words). Additionally, we use 1B item titles from our e-commerce platform. Each item title contains, on average, ten words (10B words). The latter means that three words out of four words belong to our e-commerce data during training (10B/3.3B~3).
We provide a few examples of item titles in Table 1 (b)). 
The 1B item titles are grouped into separate documents by their leaf category (17k leaf categories in total). Leaf categories are finely granular; for example, "Pencil Sharpeners", "Restaurant Bar Stools", "Cold Beverage Dispensers", "Cell Phone and Smartphone Parts", "Tuxedo and Formal Shirts", "Wallets and Coin Purses", "Can Opener Parts and Accessories", etc. Since we group the titles by their categories into separate files, the original BERT Next Sentence Prediction (NSP) task operates as a category prediction task. 
We observed that the category prediction task already attains high accuracy (over 90 percent) in the early stages of the training procedure and goes up to 99 percent, compared to the BooksCorpus and English Wikipedia data, which reach 98 percent.
We use WordPiece \cite{wu2016googles} as in the original BERT paper with a token vocabulary of size 50k instead of 30k, considering 
our unique e-commerce corpus. 

After pre-training the eBERT model, we fine-tune it for the QTR binary classification task, which outperforms the original BERT model and therefore serves as our teacher.
The input for the fine-tune task is a concatenation of query and title tokens with the following structure: 
\begin{equation}
   \mathtt{[CLS] query... [SEP] title... [SEP]}
   \label{eq:concatinput}
\end{equation}
We prune the longer input between the query and title until the concatenation has at most 23 tokens (including the SEP and CLS tokens).
At its CLS output, BERT computes a feature vector of size $768$. We create a classifier for the QTR task by adding a softmax layer on top.
We jointly train the parameters of BERT and the classifier during fine-tuning by maximizing the probability of the correct label, using the cross-entropy loss.
\subsection{Student Architectures}
\label{sec:students}

\begin{figure}
  \centering
  \includegraphics[width=0.32\textwidth]{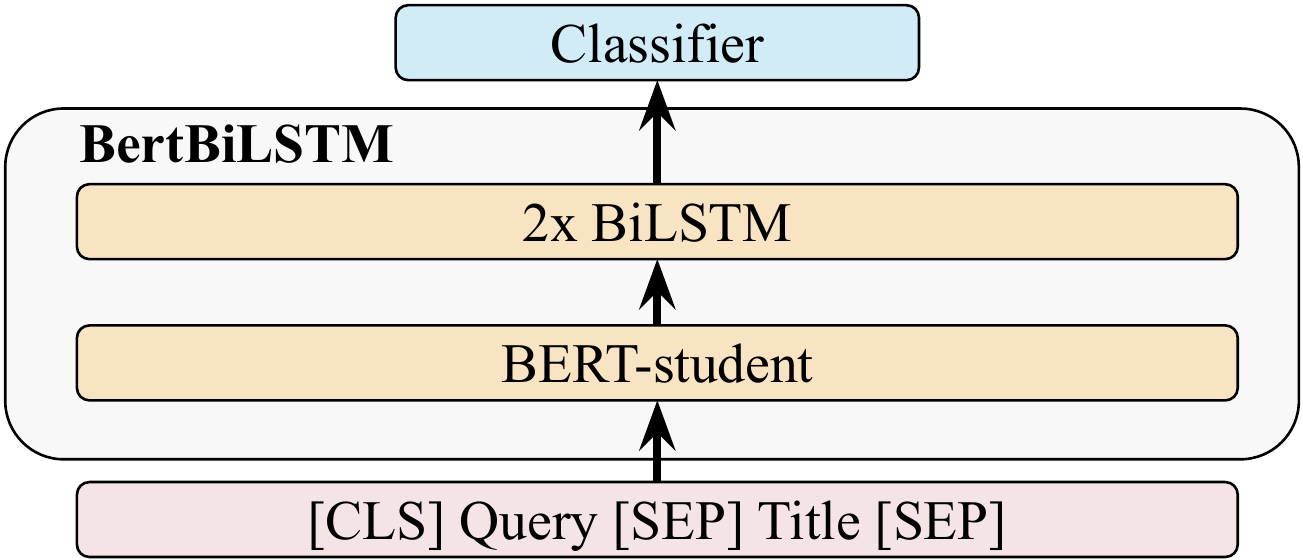}
  \caption{The BertBiLSTM model.}
  \label{fig:bertbilstm}
\end{figure} 



Due to the tight constraints on latency and throughput in production, we investigate KD to train significantly faster and smaller student models that learn from our fine-tuned eBERT model. Our goal is to capture the quality of the eBERT teacher, while keeping the student model small and efficient.
We follow the architecture blueprint proposed in \cite{tang2019distilling} for task-specific KD with BiLSTMs, but additionally consider various encoder architectures and depths. In order to reduce the architectural search space, we only investigate models in which all layers have the same hidden size $d_\text{h}$. As its input, the models receive the concatenation of query and title as described in Equation \ref{eq:concatinput}.\footnote{In \cite{tang2019distilling}, it is suggested to use a siamese BiLSTM for tasks with sentence-pair input. However, we found that this performs worse than using a single encoder for our task. This is shown for BiLSTMs with $d_\text{h}=300$ in Table \ref{tab:model_overview}.} The embedded input is run through the encoder, the output vector of which is then passed through a ReLU-activated hidden layer with output size $d_\text{h}$. Finally, the state is projected to the binary output logits.



We first investigate encoders consisting purely of BiLSTMs or BERT-student layers.\footnote{When BERT is the last encoder layer, we apply mean pooling to determine the output vector. In our experience, this works better than taking the CLS output for shallow models.} Next, we hypothesize that BiLSTM and BERT have complementary advantages in our setting, motivating us to combine the two architectures in a hybrid encoder. The model BertBiLSTM consists of a single BERT-student layer on the bottom, whose position-wise states are then fed into one or two BiLSTM layers as shown in Figure \ref{fig:bertbilstm}.


Even with our highly-optimized CPU inference toolkit, we need to rely on extremely small models in order to meet the inference latency constraint of 0.2ms per query-title input.
Guided by Table \ref{tab:speed}, we can either use a single BiLSTM or BERT-student encoder layer with hidden size $d_\text{h}=300$, or up to three encoder layers (BERT-student and/or BiLSTM) with hidden size $d_\text{h}=128$. For BERT-student layers, we always set the intermediate size $d_\text{ff}=d_\text{h}$ and the number of heads $h=4$. 
In \cite{tang2019distilling}, word2vec \cite{mikolov2013efficient} is used for the student's token embeddings. We do not have word2vec embeddings readily available for our domain, so instead we rely on the embedding table learned by the fine-tuned eBERT teacher model. We apply a linear layer to the embeddings, reducing their size from 768 to $d_\text{h}$, in order to match the encoder input size. Parameters of this linear layer are learned jointly with the rest of the student model, whereas the embedding table is not updated.
To accommodate self-attention layers, we add token type embedding and positional embedding to the encoder input  as in \cite{devlin2018bert}.

\section{Experiments}
In this section, we present the results of the fine-tuned BERT teacher models and the significantly faster and smaller student models on our QTR task. We evaluate model performance with the ROC AUC score \cite{fawcett2006introduction}, the average precision \cite{manning2008introduction} (AvgPrec) and cross entropy (CE). 
We train the student models with the AdamW optimizer \cite{loshchilov2017decoupled}
\footnote{Although the BiLSTM distillation work in \cite{tang2019distilling} uses AdaDelta \cite{zeiler2012adadelta}, we have found that AdamW \cite{loshchilov2017decoupled} performed better on our QTR task.}
for at most 100 epochs and terminate training early if ROC AUC on the development set does not improve for 30 epochs. The test set results are given for the model checkpoint with the best performance on the development set. Following \cite{tang2019distilling}, we set the batch size to 256. For each student model architecture, we run separate trainings with learning rates $10^{-3}$, $0.5\cdot10^{-4}$, and $10^{-4}$ and select the model with the highest ROC AUC score on the development set.
\footnote{In preliminary experiments, we considered using dropout with dropout rates of 0.1, 0.2, and 0.3. Since the models are already small for the QTR task, and dropout serves as a regularization method, we found that it only worsened the student performance. Therefore, dropout is disabled for all student models presented in this paper.}


Data augmentation can be an important component of KD \cite{tang2019distilling}, since the amount of labeled
data available for downstream tasks is often not sufficient to properly train student models.
With data augmentation, we can alleviate this issue
by generating plausible unlabeled data, and then computing target logits with the teacher model.
We follow \cite{tang2019natural} for language model based data augmentation.\footnote{On the QTR task, generation with language models outperforms rule-based data augmentation \cite{tang2019distilling} by a large margin.} We fine-tune GPT2-medium \cite{radford2019language} on the QTR data as structured in Equation \ref{eq:concatinput} for one epoch, using a learning rate of $5\cdot10^{-5}$ (with linear decay) for the AdamW optimizer \cite{loshchilov2017decoupled} and a batch size of 16. Then, we sample augmented data from the language model, obtaining 30.1M examples after filtering for overlap.\footnote{In order to not contaminate our training data, we remove any generated example whose query or title is also contained in the development or test set.} In most experiments, we train students on this data, concatenated with the original 764k labeled examples.


\subsection{Teacher Model}
We have fine-tuned both the original BERT-base model \cite{devlin2018bert} as well as our eBERT-base model on the QTR task. eBERT has been pre-trained on additional e-commerce data, as explained in Section \ref{sec:teacher}. The use of domain-specific data provides an enormous boost to model performance over BERT-base, as we can see in Table \ref{tab:model_overview}. 
We use the fine-tuned eBERT-base model as the teacher for all experiments with KD.

\subsection{Student Models}
\begin{table}[!t]
\footnotesize
\renewcommand{\tabcolsep}{3.8pt}

    \centering
    \caption{The table summarizes teacher and student model performances on the QTR test set. The fine-tuned eBERT-base model serves as a teacher to train the small student models (BiLSTM, BERT-student, or our combination of the two in BertBiLSTM; S-BiLSTM refers to the siamese BiLSTM as described in \cite{tang2019distilling}). $L$ is the number of encoder layers, $d_\text{h}$ the hidden size, and \#params the number of trainable parameters.}
    \begin{tabular}{lllrrrr}
         \toprule
         Model & L & $d_\text{h}$ & ROC AUC & AvgPrec &  CE  & \#params \\ \midrule
         BERT-base \cite{devlin2018bert} & 12 & 768 & 82.26 & 93.91 & 0.3837 & 109.5M \\
         eBERT-base & 12 &  768 & \textbf{87.05} & \textbf{95.60} & \textbf{0.3362} & 124.2M \\ 
         \midrule
         BERT-student & 1 & 128 & 84.83 & 94.87 &  0.3577 &  0.3M \\
         BiLSTM \cite{tang2019distilling} & 1 & 128 & 84.82 & 94.88 &  0.3574 &  0.9M \\
         BERT-student & 1 & 300 & 85.74 & 95.20 &  0.3488 &  1.0M \\
         BiLSTM \cite{tang2019distilling} & 1 & 300 & 85.76 & 95.17 &  0.3476 &  2.0M \\ \midrule
         BERT-student & 2 & 128 & 85.73 & 95.22 &  0.3474 &  0.4M \\
         BiLSTM & 2 & 128 & 85.65 & 95.17 &  0.3490 &  0.9M \\
         BertBiLSTM & 2 & 128 & 86.13 & 95.34 &  0.3433 & 0.6M \\
         \midrule
         BERT-student & 3 & 128 & 86.13 & 95.33 &  0.3442 & 0.5M \\
         BiLSTM & 3 & 128 & 85.94 & 95.27 &  0.3455 & 1.3M \\
         BertBiLSTM & 3 & 128 & \textbf{86.30} & \textbf{95.40} &  \textbf{0.3426} & 1.0M \\
         \midrule
         S-BiLSTM \cite{tang2019distilling} & 1 & 300 & 85.38 & 95.08 &  0.3522 & 2.6M \\
         \bottomrule
    \end{tabular}
    \label{tab:model_overview}
\end{table}
\begin{figure}[t]
  \centering
  \input{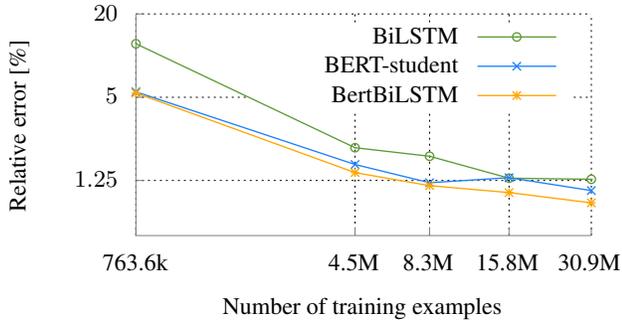}
  \caption{The figure provides the error of three-layer student models' ROC AUC, relative to the teacher, on the test set, with different amounts of training data (original data plus augmented data \cite{tang2019natural}). Both axes are on logarithmic scale.}
  \label{fig:augmentation}
\end{figure} 

\begin{table}[!t]
\footnotesize
\renewcommand{\tabcolsep}{3.8pt}

    \centering
    \caption{Scores on the test set (ROC AUC) when training the small student models without data augmentation. We compare results with and without Knowledge Distillation (KD).}
    \begin{tabular}{lllrrr}
         \toprule
         Model & L & $d_\text{h}$ & KD & No KD &   \#params \\ \midrule
         BERT-student & 1 & 128 & 78.70 & 74.67 &    0.3M \\
         BiLSTM \cite{tang2019distilling} & 1 & 128 & 75.01 & 72.89 &    0.9M \\
         BERT-student & 1 & 300 & 78.68 & 72.34 &    1.0M \\
         BiLSTM \cite{tang2019distilling} & 1 & 300 & 75.02 & 72.25 &    2.0M \\ \midrule
         BERT-student & 2 & 128 & 82.58 & 75.89 &    0.4M \\
         BiLSTM & 2 & 128 & 75.36 & 73.21 &    0.9M \\
         BertBiLSTM & 2 & 128 & 81.92 & 74.32 &   0.6M \\
         \midrule
         BERT-student & 3 & 128 & 82.29 &  78.74 &   0.5M \\
         BiLSTM & 3 & 128 & 74.96 & 73.00 &   1.3M \\
         BertBiLSTM & 3 & 128 & 82.38 & 74.29 &   1.0M \\
         \bottomrule
    \end{tabular}
    \label{tab:model_overview_hard_targets}
\end{table}


Table \ref{tab:model_overview} shows the results that we obtained with different model architectures. All student models outperform the off-the-shelf BERT-base while being at least 80x faster in inference. Most of the student models even come impressively close to the quality of the teacher model eBERT-base. Note that, for our use-case, every further improvement is essential since it can lead to better search results. Comparing the various student architectures, we observe that BiLSTM and BERT-student have almost identical performance in one-layer models. Still, BERT-student performs better in deeper models, even though it has fewer parameters. Generally, we see that deep and narrow models work better than the shallow variants. The hybrid BertBiLSTM improves over pure encoders by 0.4 ROC AUC for two-layer models and 0.17 ROC AUC (0.27 on the development set) for three-layer models. Interestingly, the two-layer BertBiLSTM even performs similarly to the best pure three-layer encoders. These results show that the hybrid model has merits in low-latency deployment scenarios.

Next, we investigate the impact of data augmentation, varying the amount of data added to the original 764k training examples between zero and 30.1M. The results, shown in Figure \ref{fig:augmentation}, demonstrate that data augmentation is a crucial aspect of knowledge distillation for this task. Without data augmentation, the performance of all student models is much worse than that of the teacher. As we add more augmented data, the students' performance comes close to the teacher. Pure BiLSTM models require significantly more training data than BERT-student to achieve comparable quality. Starting with more than 8.3M training examples, the hybrid encoder architecture BertBiLSTM consistently outperforms the other student models.

The work in \cite{rebuttal_reviewer_2} has shown that KD is more effective for multi-class problems (where logits contain information about the relationship between different classes) and that KD for binary tasks can be interpreted as example re-weighting (as described in Proposition 1 in \cite{rebuttal_reviewer_2}). To reveal the importance of KD training in our application, we train the small models provided in Table 3 with hard targets. We can see from Table 4 that, for all models, KD with soft targets outperforms training with cross-entropy towards the hard ground-truth labels. As previously mentioned, another significant advantage of KD is that we can enlarge our training data via data augmentation since we can generate additional logits with the teacher. We can see the positive effect of data augmentation in Figure 2. Due to data augmentation, the results in Table 3 consistently outperform those in Table 4, an outcome that is attainable in the first place due to the KD usage.

\section{Conclusions}
Inference with the BERT model is computationally expensive and extremely challenging for deployment in real-world production systems. We significantly outperform the off-the-shelf BERT model for the QTR classification task while at the same time making it feasible for deployment. We achieved this result by leveraging our unique data to create the eBERT teacher model and then compressing it, using Knowledge Distillation, into a compact student architecture, which we denote as BertBiLSTM.  The proposed model combines the benefits of a minuscule BERT model with a BiLSTM layer and can be easily deployed in our production systems to boost our search engine performance. The proposed architecture also outperforms other student models trained with Knowledge Distillation.

\bibliographystyle{IEEEbib}


\section{Acknowledgements}
We would like to thank Michael Kozielski for building the pre-trained eBERT model, Shubhangi Tandon for the fine-tuned teacher models, and Pavel Petrushkov for inference speed results.

\bibliography{refs}


\end{document}